# AdaBoost-assisted Extreme Learning Machine for Efficient Online Sequential Classification

Yi-Ta Chen, Yu-Chuan Chuang, and An-Yeu (Andy) Wu, *Fellow, IEEE*
Graduate Institute of Electronics Engineering, National Taiwan University, Taipei, Taiwan
{edan, frankchuang}@access.ee.ntu.edu.tw, andywu@ntu.edu.tw

*Abstract*— In this paper, we propose an AdaBoost-assisted extreme learning machine for efficient online sequential classification (AOS-ELM). In order to achieve better accuracy in online sequential learning scenarios, we utilize the cost-sensitive algorithm-AdaBoost, which diversifying the weak classifiers, and adding the forgetting mechanism, which stabilizing the performance during the training procedure. Hence, AOS-ELM adapts better to sequentially arrived data compared with other voting based methods. The experiment results show AOS-ELM can achieve 94.41% accuracy on MNIST dataset, which is the theoretical accuracy bound performed by original batch learning algorithm, AdaBoost-ELM. Moreover, with the forgetting mechanism, the standard deviation of accuracy during the online sequential learning process is reduced to 8.26x.

## I. INTRODUCTION

Extreme learning machine (ELM) is a single layer feedforward neural network (SLFN) proposed by Huang *at el.* as shown in Fig. 1(a) [1]. Different from a traditional neural network trained by back propagation (BP), input weights of ELM are randomly assigned and output weights of ELM are calculated by Moore-Penrose pseudo inverse. Due to the hardware-feasible training process, ELM has been proved to have fast training speed and good generalization performance [2].

However, in many scenarios, training data arrives sequentially and the model needs to be updated real time, which will make the batch learning ELM algorithm not efficient due to delay time of collecting all training data. Fortunately, Liang *at el.* [3] proposed an online sequential extreme learning machine (OS-ELM), which could learn data one-by-one or chunk-by-chunk.

In addition, to achieve better performance and stability in OS-ELM, ensemble is a commonly used technique by concatenating several weak classifiers into a strong classifier. Lan *at el.* [4] proposed an ensemble based OS-ELM (EOS-ELM) which adds up the results of weak classifiers to form the final output. Cao *at el.* [5] proposed a voting based online sequential extreme learning machine (VOS-ELM) whose final outputs were decided by majority voting among all classifiers. Mirza *at el.* [6] proposed voting based weighted OS-ELM (VWOS-ELM) which has a cost factor for each input data to tackle class imbalance problems.

This research was supported in part by the Ministry of Science and Technology of Taiwan (MOST 106-2221-E-002-205-MY3).

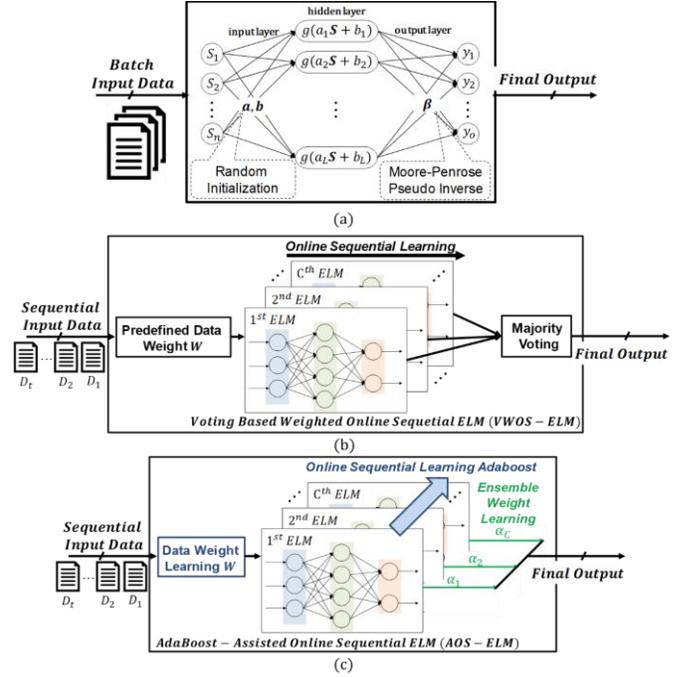

Fig. 1. Overview of (a) single extreme learning machine, (b) traditional voting based weighted online sequential extreme learning machine, and (c) proposed AdaBoost-assisted online sequential extreme learning machine.

Although VWOS-ELM has improved the performance and the stability of ELM, there are still some problems. First, VWOS-ELM is designed to conquer the class imbalance problem, which is usually a binary task. Second, VWOS-ELM rarely managed to train weak classifiers with misclassified samples, which might be predicted correctly by using larger cost factors. Moreover, several cost-sensitive algorithms have been proposed to enhance the performance of the classifier, such as RealBoost [7], LogitBoost [7], and AdaBoost [8]. However, these algorithms are not designed to be used in online sequential learning scenarios. Simply applying these algorithms may cause unstable effects on the performance.

In this paper, we propose an AdaBoost-assisted online sequential ELM (AOS-ELM) to conquer the problems above. Different from previous mentioned ELMs, the proposed AOS-ELM has a cost value for each input data, which will be updated in the training process with AdaBoost algorithm. Moreover, we calculate the final ensemble weight according to the present performance and the previous weight of the classifier using the concept of the forgetting mechanism. As shown in Fig. 1, our

main contributions are as follows:
1) **Proposed AdaBoost-assisted online sequential ELM**: Inspired by OS-ELM and AdaBoost, when data is trained through each classifier, if the prediction from previous classifier is wrong, its cost will increase. Therefore, the next classifier will try to predict the data correctly since it has a larger cost. The experiment results show that the AOS-ELM can hit the closest accuracy which is **94.41%** to the theoretical accuracy bound performed by original batch learning algorithm, AdaBoost-ELM [12].
2) **Ensemble weight calculation with forgetting mechanism**: In order to make performance more stable in the online sequential learning phase, the final ensemble weight of each classifier is decided according to not only the accuracy of the present chunk of data but also the previous ensemble weight of the classifier, which is the trade-off between temporal and spatial. The ratio between the factor of present performance and the previous ensemble weight is the forgetting factor. The standard deviation of AOS-ELM can be reduced to **8.26x** than that without a forgetting mechanism.

## II. BACKGROUND AND RELATED WORKS

### A. Extreme Learning Machine (ELM), Online Sequential ELM (OS-ELM), and Weighted OS-ELM (WOS-ELM)

For a multiclass classification problem, assumed there are $M$ classes and each data has $D$ dimensions. The input dataset $\aleph$ can be denoted as $\{(x_i, t_i)\}_{i=1}^{N}$, where $x_i \in \mathbb{R}^{1 \times D}$ is the input data vector, $t_i \in \mathbb{R}^{1 \times M}$ is the target vector for each input data and $N$ is the size of input dataset. If $g(x)$ is an activation function of the hidden layer, such as sigmoid function or ReLU function, outputs of an ELM can be denoted as

$$o_i = \sum_{j=1}^{L} \beta_j\, g(a_j, b_j, x_i), \ i=1,\dots,N, \qquad (1)$$

where $a_j$ and $b_j$ are randomly assigned input weights and biases of hidden layers, $L$ is the number of hidden nodes, and $\beta_j$ is the output weight vector which connects the hidden node $j$ to the output layer. The equation (1) can be presented as a matrix form

$$O = \beta H, \qquad (2)$$

where $H \in \mathbb{R}^{N \times L}$ is the matrix of hidden layer output

$$H = \begin{bmatrix} g(x_1 \cdot a_1 + b_1) & \cdots & g(x_1 \cdot a_L + b_L) \\ \vdots & \ddots & \vdots \\ g(x_N \cdot a_1 + b_1) & \cdots & g(x_N \cdot a_L + b_L) \end{bmatrix}. \qquad (3)$$

In order to solve the $\beta$ of equation (2), we can apply the least mean square solution to get

$$\beta = H^\dagger T = (H^T H)^{-1} H^T T, \qquad (4)$$

where $H^\dagger$ is the Moore-Penrose generalized inverse of matrix $H$ and $T$ is the matrix form of all the target vector $t_i$.

OS-ELM has been proposed based on the aforementioned ELM theory. Assumed the initial input data is $\aleph_0 = \{(x_i, t_i)\}_{i=1}^{N_0}$, where $N_0$ is the size of the initial data, and an upcoming chunk can be denoted as $\aleph_t = \{(x_i, t_i)\}_{i=\sum_{j=1}^{t-1} N_j+1}^{\sum_{j=1}^{t} N_j}$, where $N_i$ is the size of chunk at time $i$. The training process of OS-ELM can be separated into two steps:

**Step1:** *Initialization*

Given an initial batch of input data $\aleph_0$, and the random input weight $a_j$ and bias $b_j$. The initialization of the output weight can be denoted as

$$\beta^{(0)} = P^{(0)} H_0^T T_0, \qquad (5)$$

where

$$P^{(0)} = (H_0^T H_0)^{-1}, \qquad (6)$$

and $H_0$ is the initial output of the hidden layer.

**Step2:** *Online Sequential Learning*

In the online learning phase, given a new arrival of input data $\aleph_{t+1}$ in time $t+1$, the output weight can be updated as

$$\beta^{(t+1)} = \beta^{(t)} + P^{(t+1)} H_{t+1}^T (T_t - H_{t+1} \beta^{(t)}), \qquad (7)$$

where

$$P^{(t+1)} = P^{(t)} - P^{(t)} H_{t+1}^T (I + H_{t+1} P^{(t)} H_{t+1}^T)^{-1} H_{t+1} P^{(t)}, \qquad (8)$$

and $H_{t+1}$ is the output of hidden layers of $(t+1)th$ chunk of data.

WOS-ELM is proposed to embed the cost of each data into the OS-ELM training process [9], which is originally used to deal with class imbalance problems. The difference between OS-ELM and WOS-ELM is whether each input data has its own cost $\omega$. In WOS-ELM, the cost of minority class is $\omega^- = \frac{1}{m^-}$, where $m^-$ is the number of minority class samples, and the cost of majority class samples is $\omega^+ = \frac{1}{m^+}$, where $m^+$ is the number of majority class samples. The cost matrix of a chunk of data $\aleph_t = \{(x_i, t_i)\}_{i=\sum_{j=1}^{t-1} N_j+1}^{\sum_{j=1}^{t} N_j}$ can be written as

$$W_t = \begin{bmatrix} \omega_1 & \cdots & 0 \\ \vdots & \ddots & \vdots \\ 0 & \cdots & \omega_{N_t} \end{bmatrix}, \qquad (9)$$

which is the diagonal matrix of cost vector of all data

$$w_t = [\omega_1, \omega_2, \dots, \omega_{N_t}]. \qquad (10)$$

The training process of WOS-ELM can also be separated into two steps as same as OS-ELM but with some minor modifications as following:

**Step1:** *Initialization*

The initialization of output weights can be denoted as

$$\beta^{(0)} = P^{(0)} H_0^T W_0 T_0, \qquad (11)$$

where

$$P^{(0)} = (H_0^T W_0 H_0)^{-1}. \qquad (12)$$

**Step2:** *Online Sequential Learning*

In the online learning phase, the output weight can be updated as

$$\beta^{(t+1)} = \beta^{(t)} + P^{(t+1)} H_{t+1}^T W_{t+1} (T_t - H_{t+1} \beta^{(t)}), \qquad (13)$$

where

$$P^{(t+1)} = P^{(t)} - P^{(t)}H_{t+1}^T(W_{t+1}^{-1} + H_{t+1}P^{(t)}H_{t+1}^T)^{-1}H_{t+1}P^{(t)}. \quad (14)$$

*B. AdaBoost Algorithm*

AdaBoost algorithm [10] is one of the well-known ensemble methods, which combines a series of sequentially trained classifiers to form a stronger model. To increase the diversity among classifiers, each classifier is trained by same data but each of them has different weights in cost function. The weight of each data will be scaled up or down when a classifier misclassifies the data or not.

After training all weak classifiers, a final decision is made by using an ensemble weight $\alpha$ of each classifier decided by the weighted training accuracy. Higher weighted training accuracy will lead to a larger $\alpha$ that is more significant to the final output. The final output vector of the ensemble model can be computed as

$$o_{final} = \sum_{c=1}^{C}\alpha_i o_i, \quad (15)$$

where $c = 1, \ldots, C$ and $C$ is the number of classifiers. Equation (15) is the element-wise weighted sum of the output vectors of total $C$ classifiers.

### III. PROPOSED ADABOOST-ASSISTED ONLINE SEQUENTIAL EXTREME LEARNING MACHINE

In order to achieve higher performance than the other ensemble models, we proposed an AdaBoost-assisted extreme learning machine for online sequential learning scenario. The AOS-ELM can be described in three parts as shown in Fig. 2: online sequential learning AdaBoost algorithm, ensemble weight learning with forgetting mechanism, and the inference procedure.

*A. Online Sequential Learning AdaBoost Algorithm for ELM*

Different from directly applying AdaBoost to ELM, the proposed AOS-ELM has to apply an updated procedure of the data cost matrix $W$ and the ensemble weight $\alpha$ for each classifier. If the previous classifier predicts data wrong, the cost factor of the data will increase, which will make the following classifier to increase the correctness of the data. The ensemble weight can reflect the weighted performance of each classifier. As same as WOS-ELM, the training procedure of AOS-ELM can be divided into two steps:

**Step1:** *Initialization*

Given an initial data chunk $\aleph_0 = \{(x_i, t_i)\}_{i=1}^{N_0}$, the initial cost matrix $W_0$ will be initialized as following

$$W_0 = diag(\tfrac{1}{N_0}, \tfrac{1}{N_0}, \ldots, \tfrac{1}{N_0}), \quad (16)$$

where $N_0$ is the number of initial chunk of data and $N_0$ must follow the constraint to get the initial output weight $\beta$

$$N_0 \geq L, \quad (17)$$

The input weight $a_j$ and $b_j$, where $j = 1, \ldots, L$, of the ELM are randomly assigned. After training the first ELM using equations (11)-(12), we can get the output weight $\beta$. By feeding the initial

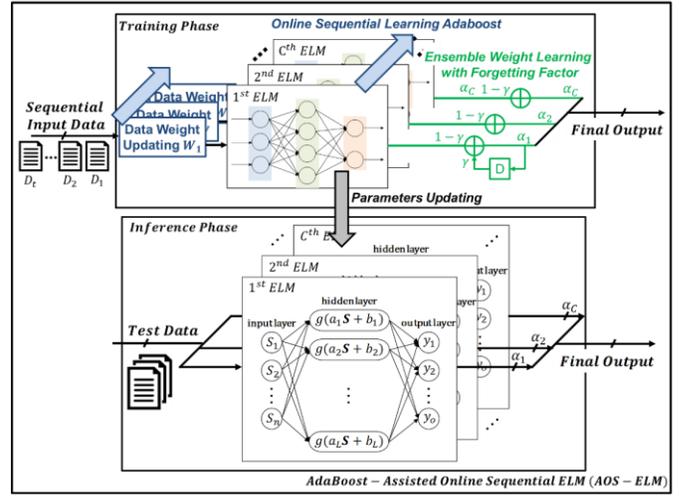

Fig. 2. Architecture of proposed AdaBoost-assisted online sequential extreme learning machine (AOS-ELM).

data $\aleph_0$ into the first ELM, we can calculate the weighted error rate of the first ELM

$$e = \sum_{n=1}^{N_0} W_0\, I(o(x_n) \neq y_n)/\sum_{n=1}^{N_0} W_0. \quad (18)$$

From AdaBoost algorithm for multiclass classification [10], we calculate the ensemble weight of the first ELM

$$\alpha = log\,(1 - e)/e + log(M - 1), \quad (19)$$

where $M$ is the number of the class. Finally, we can update the data weight for the following ELMs using the following equations

$$W_{c+1} = W_c\, exp\big(\alpha_c I(O_n(x_n) \neq y_n)\big), \quad (20)$$

$$W_{c+1} = \frac{W_{c+1}}{\sum_{n=1}^{N_0} W_{c+1}}, \quad (21)$$

where $c = 0, \ldots, C - 1$. The following classifier can be initialized using the new updated data cost matrix $W_c$. Therefore, we can calculate the parameter $\beta_c$ and $\alpha_c$ for each classifier c using equations (11)-(12), and (19).

**Step2:** *Online Sequential Learning*

After initialization of the AOS-ELM, when a new chunk of data $\aleph_t = \{(x_i, t_i)\}_{i=\sum_{j=1}^{t-1} N_j+1}^{\sum_{j=1}^{t} N_j}$ arrives, the cost matrix of the input chunk of data can be initialized as

$$W_0 = diag(\tfrac{1}{N_t}, \tfrac{1}{N_t}, \ldots, \tfrac{1}{N_t}), \quad (22)$$

where $N_t$ is the number of present data and $N_t$ must follow the constraint

$$N_t \geq 1, N_t \in \mathbb{Z}. \quad (23)$$

We can update the first ELM using the initial $W_0$, the matrix $P$ and the output weight $\beta$ by equation (13)-(14). After updating the output weight, we can calculate the weighted error rate of the first ELM by equation (18) and calculate the new ensemble weight $\alpha$ by equation (19). The updating of the cost matrix $W_c$ is calculated by equation (20)-(21). After calculating the new cost matrix $W_c$, we can update the parameter $\beta_c$ and $\alpha_c$ for each classifier $c$ afterwards.

## B. Ensemble Weight Learning with Forgetting Mechanism

Although the previous algorithm can easily utilize the benefit of AdaBoost algorithm to enhance the classifier's performance in online sequential multiclass scenarios, the original AdaBoost algorithm of deciding the ensemble weight will cause another problem. When calculating the ensemble weight $\boldsymbol{\alpha}$, we need to calculate the error $e$ of the classifier first. However, the error of each classifier is calculated only based on the present chunk. If unfortunately, the present chunk has many outliers, the performance of the classifier will be negatively affected. In order to tackle the problem, we introduce the concept of forgetting mechanism and instead of calculating $\boldsymbol{\alpha}_{t+1}$ just according to the error, we consider the previous $\boldsymbol{\alpha}_t$ at the same time.

The new ensemble weight $\boldsymbol{\alpha}$ is calculated as following

$$\alpha_{t+1} = \gamma \alpha_t + (1-\gamma)(\log(1-e)/e + \log(M-1)), \quad (24)$$

where $\gamma$ is the ratio between present data performance and the previous ensemble weight. Therefore, $\gamma$ has to follow the constraint:

$$0 \leq \gamma \leq 1, \gamma \in \mathbb{R}. \quad (25)$$

The larger the $\gamma$ is, the more dependent the ensemble weight will be on the performance of the previous data. Therefore, the improvement of the calculation of ensemble weight $\boldsymbol{\alpha}$ can contain the information of the classifier performance on all previous data. Also, for each different scenarios, we can decide a best $\gamma$ for the calculation of the ensemble weight to avoid severe oscillation of the performance during the online sequential learning process. The overall training algorithm of AOS-ELM is summarized in Algorithm 1.

## C. Inference Procedure of AOS-ELM

After calculating all the parameters for AOS-ELM, we can predict an unseen input data in the inference phase. In the inference procedure, given an unseen data chunk $\aleph_k$, we calculate the output of the hidden layer $H$ by input weights $a_j$, bias $b_j$ and the activation function with equation (3) for every classifier. Afterward, we can calculate the output layer $o_c$ for each classifier $c$. Finally, from concatenating all the classifier results using equation (15), we can predict the final class $C_{ans}$ of the data by

$$C_{ans} = \underset{c \in \{1,\dots,M\}}{\operatorname{argmax}} \; o_{final} \quad (26)$$

## IV. EXPERIMENT RESULTS

### A. Experimental Settings for Online Sequential Classification Problem

In order to evaluate the performance of AOS-ELM, we apply a digits hand written dataset MNIST [11] to verify the performance of AOS-ELM comparing to other ensemble ELM algorithms. Moreover, we partitioned the MNIST to simulate the online multiclass scenario. In the MNIST dataset, input data is an image of size 32x32 pixels, whose dimensions are 1024. Each image is a hand-written digit range from 0-9. Therefore, classifiers are used to predict the digit of an input image, which is a multiclass classification problem. In order to lower training time and reduce the number of hidden neurons, we first reduce the dimension of an input image by using principle component analysis (PCA) to reserve 90% data variance. The final input dimension of training data is 87. The training dataset has the size of 60,000, and the testing dataset has the size of 10,000.

The final accuracy of each model is calculated by averaging among ten trials and the standard deviation is calculated using the final accuracy starting from the middle of the online learning process until the end of it. First, we evaluate the performance of AOS-ELM with different forgetting factors. After selecting the optimized value of $\gamma$, we compare the AOS-ELM with the other models. We use accuracy and standard deviation as evaluation metrics. Moreover, because performance of each trial may have some differences, we repeat 10 times for the evaluation process of each classifier, and the overall performance of the classifier is averaged.

---

**Algorithm 1** Training for AOS-ELM

**Given** $g(x), C, D, L, M$
**Input**: Initialize data $\aleph_0$, Sequentially input data $\aleph \in \{\aleph_1, \aleph_2, \dots, \aleph_t, \dots\}$.
**Output:** Trained ELM parameters $a_j, b_j, \boldsymbol{\beta}, \boldsymbol{\alpha}$.

**Initialization Phase:**
1: Generate data weight $W = diag(1/N_0, \dots, 1/N_0)$
2: **for** classifier $c$ from 1 to $C$ **do**
3:    Generate $a_j$ and $b_j$ for every hidden node $j$
4:    Compute hidden node matrix $H$
5:    Compute $P_0 = (H_0^T W_0 H_0)^{-1}$
6:    Compute $\boldsymbol{\beta}^{(0)} = P_0 H_0^T W_0 Y_0$
7:    Compute $e = \sum_{n=1}^{N_0} W I(O_u(x_n) \neq y_n) / \sum_{n=1}^{N_0} W$
8:    Compute $\alpha_c^{(0)} = \log(1-e)/e + \log(M-1)$
9:    Update $W_{c+1} = W_c \exp\left(\alpha_c^{(0)} I(O_n(x_n) \neq y_n)\right)$
10:   Normalize $W_{c+1} = W_{c+1} / \sum_{n=1}^{N_0} W_{c+1}$
11: **end for**
12: **return** trained ELM parameters $a_j, b_j, \boldsymbol{\beta}, \boldsymbol{\alpha}$

**Online Sequential Phase:**
13: Generate data weight $W = diag(1/N_t, \dots, 1/N_t)$
14: **for** classifier $c$ from 1 to $C$ **do**
15:   Compute hidden node matrix $H$
16:   Compute $P^{(t+1)} = P^{(t)} - P^{(t)} H^T (W^{-1} + H P^{(t)} H^T)^{-1} H P^{(t)}$
17:   Compute $\boldsymbol{\beta}^{(t+1)} = \boldsymbol{\beta}^{(t)} + P^{(t+1)} H^T W (T_t - H \boldsymbol{\beta}^{(t)})$
18:   Compute $e = \sum_{n=1}^{N_0} W I(O_u(x_n) \neq y_n) / \sum_{n=1}^{N_0} W$
19:   Compute $\alpha_c^{(t+1)} = \gamma \alpha_c^{(t)} + (1-\gamma)(\log(1-e)/e + \log(M-1))$
20:   Update $W_{c+1} = W_c \exp\left(\alpha_c^{(t+1)} I(O_n(x_n) \neq y_n)\right)$
21:   Normalize $W_{c+1} = W_{c+1} / \sum_{n=1}^{N_0} W_{c+1}$
22: **end for**
23: **return** final trained ELM parameters $a_j, b_j, \boldsymbol{\beta}, \boldsymbol{\alpha}$

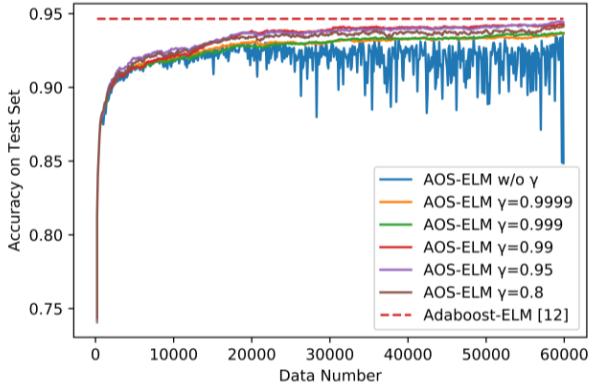

Fig. 3. Accuracy on MNIST testing set with different forgetting factor of AOS-ELM. The red line is the benchmark performance of AdaBoost-ELM.

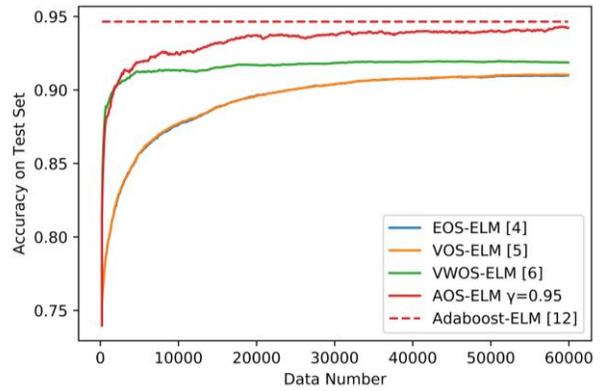

Fig. 4. Accuracy on MNIST testing set with increasing training data number. The red line is the benchmark performance of AdaBoost-ELM.

TABLE I
PERFORMANCE OF AOS-ELM WITH DIFFERENT FORGETTING FACTORS

|  | Mean Accuracy | STD of Accuracy |
|---|---|---|
| AdaBoost-ELM [12] | 94.55% | - |
| AOS-ELM, $\gamma = 0.9999$ | 93.6% | 0.0011 |
| AOS-ELM, $\gamma = 0.999$ | 93.67% | 0.0014 |
| AOS-ELM, $\gamma = 0.99$ | 94.24% | **0.0007** |
| AOS-ELM, $\gamma = 0.95$ | **94.41%** | 0.0015 |
| AOS-ELM, $\gamma = 0.8$ | 94.13% | 0.0013 |
| AOS-ELM w/o $\gamma$ | 84.86% | 0.0124 |

TABLE II
PERFORMANCE OF ALL MODELS

|  | Mean Accuracy | STD of Accuracy |
|---|---|---|
| AdaBoost-ELM [12] | 94.55% | - |
| EOS-ELM [4] | 90.98% | 0.0015 |
| VOS-ELM [5] | 91.03% | 0.0017 |
| VWOS-ELM [6] | 91.86% | **0.0003** |
| AOS-ELM w/o $\gamma$ | 84.86% | 0.0124 |
| AOS-ELM $\gamma = 0.95$ | **94.41%** | 0.0015 |

In our experiment, we focus on the comparison among the following five models: AdaBoost-ELM [12], EOS-ELM [4], VOS-ELM [5], VWOS-ELM [6], and AOS-ELM. The number of classifiers is set to 30, and the number of hidden nodes is set to 150. For online sequential algorithms, we set the initial batch size as 200, and the following chunk size as 100. For VWOS-ELM, the weight of each class data is set to the reciprocal of the number of each class data.

### B. Experiment Result of Different Forgetting Factors

Fig. 3 and Table 1 present the result of different value of forgetting factors comparing to the AOS-ELM without forgetting factor. The performance of AdaBoost-ELM is set to a benchmark according to the best performance that such size of ELM model can reach due to its batch learning characteristic. From Table 1, we can see that there is an optimized value of $\gamma$ to get the best performance. If $\gamma$ is too large, the overall performance will degraded. However, if $\gamma$ is too small, the overall performance will be unstable. Therefore, from the experiment result, we picked $\gamma$ to be 0.95 to get the highest overall accuracy which could reach 94.41% comparing to the benchmark performance of 94.55%. Moreover, by selecting the proper $\gamma$, we can lower the standard deviation by 8.26x comparing to the AOS-ELM without forgetting mechanism.

### C. Experiment Result Compared with Other Models

Fig. 4 and Table 2 present the results of five different models. As fig. 4 shown, the accuracy of AOS-ELM can reach more closely to the bounding of the baseline compared other ELM models. As shown in Table 2, AOS-ELM can get the highest final accuracy and the second lowest standard deviation among all models. Therefore, an OS-ELM utilizing the AdaBoost algorithm and forgetting mechanism can perform better than other ensemble based online ELM algorithms with same model size.

### V. CONCLUSION

In this work, we propose an AdaBoost-assisted online sequential extreme learning machine. By utilizing the learnability of AdaBoost algorithm, AOS-ELM can make the same size of ensemble ELM to reach higher accuracy in online sequential learning scenarios. By adapting different forgetting factors, AOS-ELM can reach the optimized performance and become more stable during the online sequential learning process. Therefore, the proposed AOS-ELM can outperform other ensemble based online sequential ELMs in multiclass online sequential learning scenarios.

REFERENCE

[1] G.-B. Huang, Q.-Y. Zhu and C.-K. Siew, "Extreme learning machine: theory and applications," *Neurocomputing,* vol. 70, no. 1-3, pp. 489-501, 2006.

[2] G.-B. Huang, L. Chen and C. K. Siew, "Universal approximation using incremental constructive feedforward networks with random hidden nodes," *IEEE Trans. Neural Networks,* vol. 17, no. 4, pp. 879-892, 2006.

[3] N.-Y. Liang, G.-B. Huang, P. Saratchandran and N. Sundararajan, "A fast and accurate online sequential learning algorithm for feedforward networks," *IEEE Transactions on neural networks,* vol. 17, no. 6, pp. 1411-1423, 2006.

[4] L. Yuan, Y. C. Soh and G.-B. Huang, "Ensemble of online sequential extreme learning machine," *Neurocomputing,* vol. 72, no. 13-15, pp. 3391-3395, 2009.

[5] J. Cao, Z. Lin and G.-B. Huang, "Voting base online sequential extreme learning machine for multi-class classification," in *2013 IEEE International Symposium on Circuits and Systems (ISCAS2013)*, Beijing, pp. 2327-2330, May 2013.

[6] B. Mirza, Z. Lin, J. Cao and X. Lai, "Voting based weighted online sequential extreme learning machine for imbalance multi-class classification," in *2015 IEEE International Symposium on Circuits and Systems (ISCAS)*, Lisbon, Portugal, pp. 565-568, May 2015.

[7] J. Friedman, T. Hastie and R. Tibshirani, "Additive logistic regression: a statistical view of boosting," *The annals of statistics,* vol. 28, no. 2, pp. 337-407, 2000.

[8] Y. Freund and R. E. Schapire, "A decision-theoretic generalization of on-line learning and an application to boosting," *Journal of computer and system sciences,* vol. 55, no. 1, pp. 119-139, 1997.

[9] B. Mirza, Z. Lin and K.-A. Toh, "Weighted online sequential extreme learning machine for class imbalance learning," *Neural processing letters,* vol. 38, no. 3, pp. 465-486, 2013.

[10] T. Hastie, S. Rosset, J. Zhu and H. Zou, "Multi-class adaboost," *Statistics and its Interface,* vol. 2, no. 3, pp. 349-360, 2009.

[11] Y. LeCun, L. Bottou, Y. Bengio and P. Haffner, "Gradient-based learning applied to document recognition," *Proceedings of the IEEE,* vol. 86, no. 11, pp. 2278-2324, 1998.

[12] A. Riccardi, F. Fernández-Navarro and S. Carloni, "Cost-sensitive AdaBoost algorithm for ordinal regression based on extreme learning machine," *IEEE transactions on cybernetics,* vol. 44, no. 10, pp. 1898-1909, 2014.